\documentclass{article}
\usepackage{ijcai22}

\usepackage{times}
\usepackage{soul}
\usepackage{url}
\usepackage[hidelinks]{hyperref}
\usepackage[utf8]{inputenc}
\usepackage[small]{caption}
\usepackage{graphicx}
\usepackage{amsmath}
\usepackage{amsthm}
\usepackage{booktabs}
\usepackage{algorithm}
\usepackage{algorithmic}
\usepackage{tabularx}
\usepackage{booktabs}

\title{Assisting Decision Making in Scholarly Peer Review:\\ A Preference Learning Perspective}

\author{
    Nils Dycke,\textsuperscript{\rm 1} 
    Edwin Simpson,\textsuperscript{\rm 2} 
    Ilia Kuznetsov,\textsuperscript{\rm 1}
    Iryna Gurevych\textsuperscript{\rm 1}\\
    \affiliations
    \textsuperscript{\rm 1} Ubiquitous Knowledge Processing Lab (UKP-TUDA)\\
    Department of Computer Science, Technische Universität Darmstadt.\\
    ukp.informatik.tu-darmstadt.de \\
    \emails
    \textsuperscript{\rm 2}Intelligent Systems Lab\\
    University of Bristol\\
    edwin.simpson@bristol.ac.uk
}

\newcolumntype{L}[1]{>{\raggedright\arraybackslash}m{#1}}
\newcolumntype{C}[1]{>{\centering\arraybackslash}m{#1}}
\newcolumntype{R}[1]{>{\raggedleft\arraybackslash}m{#1}}

\begin{document}

\maketitle

\begin{abstract}
Peer review is the primary means of quality control in academia; as an outcome of a peer review process, program and area chairs make acceptance decisions for each paper based on the review reports and scores they received. Quality of scientific work is multi-faceted; coupled with the subjectivity of reviewing, this makes final decision making difficult and time-consuming. To support this final step of peer review, we formalize it as a paper ranking problem. We introduce a novel, multi-faceted generic evaluation framework for ranking submissions based on peer reviews that takes into account effectiveness, efficiency and fairness. We propose a preference learning perspective on the task that considers both review texts and scores to alleviate the inevitable bias and noise in reviews. Our experiments on peer review data from the ACL 2018 conference demonstrate the superiority of our preference-learning-based approach over baselines and prior work, while highlighting the importance of using both review texts and scores to rank submissions.
\end{abstract}

\section{Introduction}
Peer review (PR) is the prevalent quality assurance mechanism in science \cite{birukou_alternatives_2011}. During peer review, multiple referees evaluate a paper providing scores and a textual report. Next, the program or area chairs weigh the reviews of all papers to make final acceptance decisions selecting the top papers exceeding a quality threshold \cite{jefferson_measuring_2002}. This requires careful comparison of papers based on the reviewers’ assessment of soundness,  presentation, impact potential and rightfulness \cite{jefferson_measuring_2002,aksnes_citations_2019}. The difficulty of this process is amplified by noise, bias and high disagreement in the peer reviews \cite{lee_bias_2013,walker_emerging_2015} and the ever-growing submission numbers in academia.

To cope with the complexity of this task, decision makers settle for heuristics to guide and prioritize the assessment process. Typically, statistics on review scores, like the mean overall score per paper, are used to rank papers as an input. Hereby, decision makers can focus on borderline regions close to the quality threshold. This heuristic is insufficient due to inconsistent usages of rating scales by the referees \cite{wang_your_2019,lee_commensuration_2015} and the aforementioned issues of noise. Consequently, agreement on absolute scores is considerably below substantial (in computer science) \cite{ragone_peer_2013} making the mean on typically three to four peer reviews per paper highly unreliable.

Peer review can gain in accuracy and efficiency when providing a reliable input ranking to the complex and time-consuming decision making task. In the past, so-called consensus ranking has been applied to rank submissions \cite{cook_creating_2007,baskin_preference_2009}; yet these approaches neglect textual review information and are prone to noise in review scores. Related methods from the field of Natural Language Processing (NLP) use textual information, like a paper’s abstract, to predict binary acceptance decisions \cite{kang_dataset_2018} or citation counts \cite{li_neural_2019}. The output of these methods is  coarse-grained or relies on information that is not available during the time of review restricting their applicability to assist acceptance decision making. 

In this paper, we propose preference learning to rank submissions represented by their review scores and texts using referees’ preferences as supervision on this feature space. According to  this perspective, training requires no external paper quality estimates, like citation counts, which are typically not available during the time of review. We investigate three hypotheses: First, preferences expressed by human referees over the set of papers under review serve as a valuable supervision signal. Second, a ranking model on submissions benefits from including peer review texts in addition to scores. Third, preference learning techniques can effectively mitigate the impact of noise, disagreement and bias in peer review data. To formalize our new preference learning perspective, we introduce a more general formulation of the assistance task, which we call the Paper Ranking Problem (PRP). To validate our hypotheses on real data, we define a novel generic evaluation framework motivated by science-of-science literature on peer review, using past acceptance labels and citation counts as a reference. We select Gaussian process preference learning (GPPL) \cite{simpson_finding_2018} to tackle the PRP. GPPL is a preference learning method that has been applied to NLP tasks, such as ranking arguments by convincingness \cite{simpson_finding_2018}, and proved to be robust against noise.

We apply the aforementioned framework on the highly structured peer review data from the 2018 conference of the Association for Computational Linguistics (ACL) \cite{gao_does_2019}. We show that our preference-learning-based approach achieves the best balanced performance on both paper quality proxies compared to previous methods and baselines. During our ablation study, we find that  review texts increase ranking performance by citation counts substantially, while review scores alone have small predictive validity for the future impact of accepted submissions. Finally, we find that our approach is less susceptible to additional noise in the review scores caused by unreliable referees and bias induced by heterogeneous weighing of quality criteria.  
\section{Related Work}
\citeauthor{cook_creating_2007} \shortcite{cook_creating_2007} propose the conversion from exact review scores to \textit{partial rankings} on a per-referee basis to mitigate the issue of score miscalibration bias in PR \cite{wang_your_2019}. For example, if referee $A$ reviewed the papers $x$, $y$ and $z$ assigning  scores of $2$, $1$ and $3$ (higher is better), respectively, this is expressed by the ordering $\prec_{A}$: $y \prec_{A} x \prec_{A} z$. To aggregate reviewed submissions into a ranking, the authors cast the task as an NP-hard consensus ranking problem: given the partial rankings on overlapping paper subsets, the goal is an output ranking on all papers that violates the least precedence pairs. The authors propose a branch-and-bound algorithm that can find optima on small artificial datasets in reasonable time. \citeauthor{baskin_preference_2009} \shortcite{baskin_preference_2009} propose a more efficient neighborhood-based optimization algorithm for the same task.
During our experiments on a real-world dataset, we show that both consensus ranking approaches perform inferior in the presence of bias and noise compared to our preference-learning-based method.

Scholarly document quality assessment (SDQA) covers various problem settings from NLP judging paper quality aspects based on paper texts or associated review texts.
\citeauthor{kang_dataset_2018} \shortcite{kang_dataset_2018} introduce the task of \textit{paper acceptance prediction} (PAP): given papers from past conferences their binary acceptance decisions are the target of prediction. This task has been approached in many flavors \cite{ghosal_deepsentipeer_2019,stappen_uncertainty-aware_2020}.
\citeauthor{maillette_de_buy_wenniger_structure-tags_2020} \shortcite{maillette_de_buy_wenniger_structure-tags_2020} combine PAP with the task of \textit{citation count prediction} (CCP), where the target of prediction is the paper's future citation count. The authors augment paper texts with structural tags and apply it on both tasks in isolation. \citeauthor{li_neural_2019} \shortcite{li_neural_2019} approach CCP using cross-attention between review texts and the paper abstract extended by author metadata like their \textit{h-index} as a feature.
Several factors hinder the applicability of SDQA methods for assistance in PR: PAP approaches make coarse-grained recommendations and often neglect review texts. Methods from CCP (and PAP) train models on historic papers from broad domains having citation counts (and acceptance decisions) available. As scientific merit is domain- and time-dependent \cite{lee_commensuration_2015}, there is a substantial distribution shift between training and inference time when applying these models for a particular, new PR process. Consequently, their recommendations are likely biased to a historic state of the art.
\section{Paper Ranking Problem}
In this section we introduce the \textit{Paper Ranking Problem} (PRP), which models the task of aggregating reviews into a ranking of submissions to assist acceptance decision making. 

\subsection{Problem Definition}
The PRP is the task of ranking the submissions to a PR system according to their estimated quality. Unlike the problem settings in PAP or CCP, the quality estimate should be produced in comparison to the other submissions, hereby accounting for their scientific context and allowing program chairs to determine the acceptance threshold dynamically.

Let $P$ denote the set of papers submitted to the venue, where each paper $p \in P$ is represented by its text. For the set of reviews $R$, each review $r \in R$ is characterized by its text and a vector of scores. Each review is associated with exactly one referee $\textit{ref}(r) \in E$ from the set of referees $E$ and exactly one paper $\textit{pap}(r) \in P$.
Hence, each referee $e \in E$ generates a set of reviews $R_e = \{r \textit{ }|\textit{  } r \in R, \textit{ref}(r) = e\}$ corresponding to a set of reviewed papers $P_e = \{\textit{pap}(r) \textit{ } | \textit{ } r \in R_e\}$.
The \textit{Paper Ranking Problem} is then defined as the task of predicting an overall ranking $\mathcal{O}_{P'}$ implying a total order on $P' \subseteq P$, given $R$, $P$ and $E$ with all associated information. $\mathcal{O}_{P'}$ should have minimal ranking distance to the ranking of papers by their true total quality ordering $\widehat{\mathcal{O}}_{P'}$.
In case of making acceptance decisions on all submissions together, we have $P' = P$. To make individual acceptance decisions on subsets of $P$ grouping submissions, for instance, by track or paper type, we model this task as multiple PRP on disjoint subsets $P'_1, ..., P'_N$ partitioning $P$. For conciseness we refer to this set of subsets by $P'$, as well.

In this definition, we assume that there is one true ranking by paper quality for each considered subset of papers. This is an inherent assumption in peer review that is merely reflected in this formalization and which is most plausible for the given context limited to papers within a single PR system. 

\subsection{Performance Criteria}
A useful assistance system increases the efficiency of a PR process, while at least maintaining the quality of output decisions. Hence, we adapt quality criteria from the science-of-science literature on peer review, to measure the performance on the PRP. Specifically, we consider \textit{effectiveness}, \textit{completeness}, \textit{fairness} and \textit{efficiency} described below.
We draw on the concept of \textit{random ranking models} \cite{critchlow_probability_1991} to describe a paper ranking algorithm $A$. The quality of a paper $p \in P'$, $u(p)$, is drawn from the true quality distribution. For any $x, y \in P'$ the pairwise precedence $x \preceq_\mathcal{M} y$ is distributed according to the random ranking model $\mathcal{M}$ implied by $A$; this means that $x$ precedes $y$ in output rankings of $A$ with the probability defined by $\mathcal{M}$.
\paragraph{Effectiveness and Completeness}
Effectiveness and completeness are classical criteria from information retrieval. In the context of PR, effectiveness is often described by the predictive validity \cite{ragone_peer_2013} or the ability to filter out low quality works \cite{birukou_alternatives_2011}. Lack of completeness is associated with low acceptance rates \cite{church_reviewing_2005}. %
If $x$ is ranked higher than $y$, effectiveness requires that the quality of $x$ is higher than the one of $y$.
\begin{equation} \label{eq:com}
    P(u(x) > u(y) | x \preceq_\mathcal{M} y) = 1
\end{equation}
For a complete ranking model, $x$ should always precede $y$ in the ranking, given that $x$ has a higher quality than $y$.
\begin{equation} \label{eq:eff}
    P(x \preceq_\mathcal{M} y | u(x) > u(y)) = 1
\end{equation}
\paragraph{Fairness}
Fairness in PR is often regarded as the absence of bias in the reviews or as the replicability of the PR process \cite{walker_emerging_2015}. This perspective neglects the subjectivity of reviewing and desirable types of bias in reviews, like a credit of trust towards potentially ground-breaking works \cite{bornmann_scientific_2011}. We propose an output-oriented criterion for fairness in the PRP: a ranking model is fair if papers of the same quality precede each other with the same probability.
\begin{equation}
    P(x \succeq_\mathcal{M} y | u(x) = u(y)) = P(y \succeq_\mathcal{M} x | u(x) = u(y))
\end{equation}
This shifts the focus of bias analysis from the reviews to the ranking model and allows for swapped pairs in the generated output rankings, as long as they do not occur systematically.
\paragraph{Efficiency}
Efficiency is a meta-criterion of the process of obtaining a ranking model. \citeauthor{ragone_peer_2013} \shortcite{ragone_peer_2013} measure the efficiency of PR by the time spent by reviewers to achieve a certain quality standard of acceptance decisions. We transfer this criterion directly to the PRP: the number of reviews required as an input should be minimal to produce an effective, complete, and fair ranking model.
\subsection{Evaluation Framework}
To evaluate approaches to the PRP in real scenarios, we approximate the latent true quality per paper $u(p)$. Being multi-faceted in nature, we combine multiple weak indicators for different aspects of quality to approximate $u(p)$.

Despite uncertainty as to what citation counts measure exactly \cite{aksnes_citations_2019}, they are common proxies for the impact of a paper. Within a fixed scientific field and a fixed time since publication, we use them as indicators for paper impact in our framework.
Furthermore, historic acceptance decisions relate to paper quality. While being venue-specific, noisy and highly selective, due to restrictive acceptance quotas, they can serve as valuable quality indicators within the context of a fixed PR process.
Although these measures should not be considered in isolation, a \textit{balanced}, high performance on both of them reveals consistency to paper quality within the limitations of the used proxies. In principle, the provided evaluation framework is applicable on any set of metrics correlating with diverse aspects of paper quality.

To measure effectiveness and completeness against a ranking by citation counts and binary acceptance decisions practically, we use Spearman's rank correlation ($\rho$) and the area under the receiver operating characteristic (AUROC), respectively.
As the probability distribution of precedence pairs $P(x \succeq_\mathcal{M} y)$ is typically not available, we approximate fairness as the sensitivity of the PRP approach to bias and noise in the input data. When adding artificial bias and noise to the review scores, a fair approach should output rankings consistent to the unaltered and true one. We hereby ground the commonly used sensitivity analysis on artificial PR data (e.g. \cite{cook_creating_2007,wang_your_2019}) on real datasets, to make it more adequate for application scenarios.
The evaluation of efficiency follows directly from its definition. After sub-sampling the set of reviews randomly, we measure the performance decrease in terms of effectiveness and completeness compared to the full dataset. During experiments we apply this general evaluation schema on real data.

\section{Preference Learning for Paper Ranking}
The Paper Ranking Problem can be naturally approached by preference learning to incorporate review texts and account for bias and noise in review scores.
\subsection{Preference-based Model of the PRP}
In preference learning, relative preferences on the item space serve as the supervision signal. For that PAP, we elicit preferences as for consensus ranking: we convert the review scores of a referee $e \in E$ into a partial ranking on the set of reviewed papers $R_e$.
This makes a latent assumption: referees consciously or subconsciously compare papers during review. While effects like \textit{order bias} \cite{birukou_alternatives_2011} suggest that this is in fact true, there might be violations for papers on very different topics reviewed by the same referee.
However, this assumption is not only induced by the preference perspective, but it is also inherent to the direct use of review scores: scores of incomparable papers are mapped to the same numeric space. In fact, the preference-based view is less restrictive, as it allows for explicit filtering of potentially invalid comparisons.

Formally, the partial rankings $\textit{PO}_E=\{\mathcal{O}_{P_e}| e \in E\}$ are given as training supervision. Here $\mathcal{O}_{P_e}$ is the total order on papers $P_e$ reviewed by $e$, which is induced by the order on associated scores. We train the model on all papers $P$, where each paper is represented by a feature vector derived from the review texts and score vectors. The goal is to predict $\mathcal{O}_{P'}$ assuming that the observed partial rankings are sampled from the true order $\widehat{\mathcal{O}}_{P'}$ with random permutations. Each partial ranking implies a set of preference pairs including \textit{tie preferences} for non-strict partial orders. In this learning setup, the model orders only items seen during training.
\subsection{GPPL for the Paper Ranking Problem}
Gaussian processes (GPs) are fully Bayesian regression learners with Gaussian priors achieving high robustness against noise in small-data domains \cite{rasmussen_gaussian_2006}.
\citeauthor{chu_preference_2005} \shortcite{chu_preference_2005} adapt GPs for preference learning. The authors assume that observed preference pairs $x \succ y$ ("$x$ is preferred over $y$"), follow a likelihood function where the probability of $x \succ y$ depends on the delta between the hidden quality of $x$ and $y$ plus Gaussian noise. GPPL predicts the quality $u(x)$ of a sample given the training pairs by marginalizing over the latent variables.

\citeauthor{simpson_finding_2018} \shortcite{simpson_finding_2018} propose a more scalable approximation for the posterior using stochastic variational inference. The authors show that scalable GPPL effectively ranks arguments by convincingness based on embeddings and linguistic features.
Scalable GPPL is suitable for our purpose given that review scores can be noisy and the number of submissions range from below hundred to thousands. Hence, we utilize the method by \citeauthor{simpson_finding_2018} \shortcite{simpson_finding_2018} for our approach.
To apply GPPL for the PRP, the partial orders $\textit{PO}_E$ are converted into a set of preferences. We enumerate all implied precedence pairs for each $\mathcal{O}_{P_e} \in \textit{PO}_E$ and join these precedence pairs of all referees into a multi-set as an input. The papers are represented by feature vectors described in the following section. The output of a GPPL model are real-valued quality estimates allowing to rank all submissions.

\subsection{Feature Set Design}
We focus on review scores and texts to represent submissions. Hereby, we investigate the importance of review texts for the PRP, and enforce that primarily the human judgements of referees are reflected in the output rankings.
We consider three feature sets to define a vector representation, such that two papers are close in the vector space if they are of similar quality.
\begin{itemize} 
    \item \textbf{Score Features}: These features are derived from the reviews' score vectors. Apart from an overall assessment, typically ratings on aspects of paper quality, like soundness or presentation, are provided as \textit{aspect scores}. 
    For each aspect score and the overall score, the mean, standard deviation, minimum and maximum over the reviews per paper are computed.
    Additionally, the score vectors of each review of a paper are concatenated in arbitrary order. Score-based features should reflect controversy and multiple quality aspects, but they are expected to be noisy, as they rely on the scores directly.
    \item \textbf{Discourse Features}: Peer reviews contain questions, feedback and summaries. The distribution of these argumentative units relates to paper quality, as they reflect its strengths and weaknesses. In the AMPERE dataset \cite{hua_argument_2019}, 400 computer science reviews are annotated with sentence-level discourse labels, including e.g. \textit{request} or \textit{non-argumentative}. We fine-tune the last layer of BERT \cite{devlin_bert_2019} for $4$ epochs on $90\%$ of samples in AMPERE resulting in a $0.7969$ micro-F1-score on the remaining $10\%$. We apply this model on each review sentence. The distribution of discourse labels and the proportion of non-argumentative sentences across the reviews of a paper are added as a feature. 
    \item \textbf{Embedding Features}: Embeddings are widely used in NLP to capture the similarity and meaning of texts. Reviews are structured into sections answering different questions of the review form. We encode each section using mean pooling on sentence-embeddings and concatenate them. Additionally, we form the mean of embeddings per section across reviews. To capture review-relatedness we compute the average cosine similarity between the first sentence per review.
    To embed sentences, we use distilled SBERT \cite{reimers-gurevych-2019-sentence} fine-tuned on the natural language inference task, as these embeddings are not domain-specific and suitable to represent general statements in reviews. In this work, we focus on a simple representation of review texts as a proof of concept. We leave domain-specific embeddings or paragraph-level embeddings for future work.
\end{itemize}
During experiments, we investigate feature subsets to determine their utility with respect to different proxies of paper quality. Additionally, we tested simple reading complexity metrics but they did not improve performance in any scenario.
\section{Experiments}
\begin{table}[t]
	\centering
	\renewcommand{\arraystretch}{0.95}
	\begin{tabular}{R{5.6cm}L{1.8cm}}
	    \toprule
		\multicolumn{2}{C{8cm}}{{\textbf{ACL-2018 Dataset Statistics}}} \\ 
		\midrule
		\# Papers & $1538$\\
		\# Reviews & $3875$\\
		\# Reviews per paper & $2.52 \pm 0.67$  \\ 
		\# Reviews per referee & $3.04 \pm 1.35$  \\ 
		Krippendorff's $\alpha$ & $0.3596$  \\
		Intra-class correlation coefficient (ICC) & $0.365$ \\
		\# Preference pairs & $5109$ \\
		\# Comparisons per paper & $6.65 \pm 2.62$ \\
		\bottomrule
	\end{tabular} 
	\caption[Basic Statistics of the Datasets]{On overall scores, we use Krippendorff's $\alpha$ \cite{krippendorff_content_1980} with an ordinal metric and ICC \cite{mcgraw1996forming}.}
	\label{t:datastats}
\end{table}
We rely on the anonymized PR data from consenting referees at ACL-2018 kindly provided by the collecting parties \cite{gao_does_2019}. According to \citeauthor{gao_does_2019}, $85\%$ of referees opted-in the collection making the dataset close to complete. ACL-2018 contains anonymous referee identifiers, as in real PR systems, which we use to infer partial rankings. To our knowledge, ACL-2018 is the only available dataset with this information.
Nevertheless, access to more complete PR data is likely in the future, due to an increased interest in open peer review \cite{birukou_alternatives_2011}.
\paragraph{Dataset} 
The ACL-2018 dataset includes reviews and acceptance labels for 1538 submissions from 21 tracks. We only consider before-rebuttal reviews to avoid social biases \cite{gao_does_2019}. The acceptance rate lies at roughly 25\%. Each review has an overall score on a six-point scale and six aspect scores  \textit{originality}, \textit{soundness}, \textit{substance}, \textit{replicability}, \textit{meaningful comparison} and \textit{readability} on five-point scales. Additionally, there are five text sections including \textit{summary and contributions}, \textit{strengths}, \textit{weaknesses}, \textit{questions} and \textit{additional comments} in each review.
Table \ref{t:datastats} summarizes relevant statistics of ACL-2018. We observe that the preference signal is ample: each paper is compared on average $6$-times. Treating PR as an annotation study, the agreement on overall scores reveals the level of consistency in referees' judgements. For ACL-2018, the agreement is low for a computer science venue \cite{ragone_peer_2013}, but high compared to social science journals \cite{bornmann_scientific_2011}.

To realize the proposed evaluation strategy on both citation counts and acceptance labels as gold standards, we match the accepted papers in ACL-2018 with the \textit{NLPScholar} dataset \cite{mohammad_examining_2020} converting the hereby acquired citation counts into a ranking. The papers have identical age making their citation counts comparable. As an additional reference, we normalize citation counts per track, which eliminates a preference to topics with broad audiences. We normalize by the sum of citation counts $\textit{cc}(p)$ of papers in track $t$ to form a ranking:
$
\textit{ncc}_\textit{t}(p) = \frac{\textit{cc}(p)}{\sum_{q \in t}\textit{cc}(q)}
$.
Due to privacy restrictions, the citation counts of rejected papers cannot be considered. 
\paragraph{Hyper-parameter Tuning and Training}
We employ the GPPL implementation by \citeauthor{simpson_finding_2018} \shortcite{simpson_finding_2018} in its default configuration with a \textit{Matérn} kernel\footnote{The code of our GPPL-based approach, baselines and detailed hyper-parameters at \url{www.anonymous-github-repo.com}}. Instead of length-scale optimization or the proposed median-heuristic we use standard normalization on the non-embedding features, as this increased performance on all gold standards substantially. We only report on variations of the feature sets, as no other hyper-parameter configuration improved performance during experiments.
The goal of the PRP is to predict the true quality ranking of papers seen during training. For all experiments, we therefore train the GPPL model on all papers and their reviews, but make predictions for paper subsets. We split ACL-2018 into a 20\% development and an 80\% test set. The sets are randomly sampled while ensuring that the distribution of acceptance labels and positions in the citation count ranking are consistent with the overall population.
We report the mean and standard deviation of the performance metrics over five runs with randomly shuffled inputs.
\paragraph{Baselines}
We consider score aggregation strategies including the mean overall score weighted by referee confidence (MEAN-S-w), median overall score (MEDIAN-S) and majority voting on overall scores (MAJOR-S) falling back to the mean score for tied votes.
The MEAN-S-w weights based on referee-provided confidence scores, as this improved performance on the development set. 
Additionally, we compare to the decision-based \cite{cook_creating_2007} (DCON) and the neighborhood-based \cite{baskin_preference_2009} (NCON) consensus rankers. Both algorithms are re-implemented in Python based on the author-provided code. They receive the same input as the GPPL model excluding tie preferences, as they cannot account for them.
\paragraph{Experimental Scenarios}
In the first experiments, we investigate the performance of different feature sets. We optimize two feature configurations, where the first is selected based on acceptance labels (by AUROC) and the second one is selected based on the citation count ranking (using Spearman's $\rho$) on the development set. Hereby, we identify which features boost performance on each gold standard. In the end, we select the configuration with the best balanced performance on both gold standards for all further experiments.

The evaluation of \textit{effectiveness} and \textit{completeness} follows directly from the generic strategy described earlier. We measure the performance of the baselines and best model on the test set of the ACL-2018 dataset.
To judge the \textit{fairness} of ranking models, we consider two scenarios of rating errors and measure their impact on  performance: First, we add random noise $\epsilon \sim \mathcal{N}(0, \sigma^2)$ with $\sigma \in \{0.75, 1.0\}$ to the aspect and overall scores (rounded to integers) of $\alpha \in \{30\%, 60\%\}$ of the referees. Hereby, we simulate the effect of unreliable referees. Commensuration bias refers to the heterogeneous weighting of paper aspects by different referees to derive the overall score of a paper \cite{lee_commensuration_2015}.
We simulate this by replacing the overall score by a weighted sum of the aspect scores and adding low normal noise ($\sigma = 0.5$). We apply this for the reviews of $\alpha = 30\%$ of the referees. To analyze different scenarios, we use equal weights (COMM-EQ), an over-emphasis on readability (COMM-READ) and discarding of the originality score (COMM-CON). %
To evaluate the \textit{efficiency} of the PRP approaches we sub-sample the reviews per paper discarding $\alpha \in \{30\%, 60\%\}$ of the reviews while guaranteeing at least one review per paper. Again we measure the performance decrease on both gold standards.
\section{Results and Analysis} \label{ss:res}

\begin{table*}[t]
	\centering
	\renewcommand{\arraystretch}{0.95}
	\begin{tabular}{R{5.2cm}C{2.5cm}C{2.5cm}C{2.5cm}C{2.5cm}}
		\toprule
		& \textbf{AUROC} & \textbf{PRAUC} & $\rho$ \textbf{raw} & $\rho$ \textbf{norm.} \\
		\midrule
		{MEAN-S-w} & $\mathbf{0.9041}$ & $0.7180$ & $0.1114$ & $0.1352$ \\
		{MEDIAN-S} & $0.8711$ & $0.6530$ & $0.1109$ & $0.1229$\\
		{MAJOR-S} & $0.8731$ & $0.6491$ & $0.1203$ & $0.1292$ \\
		{DCON} & $0.8302 \pm 0.003$ & $0.5487 \pm 0.011$ & $0.0907 \pm 0.008$ & $0.0746 \pm 0.007$ \\
		{NCON} & $0.7824 \pm 0.005$ & $0.5028 \pm 0.007$ & $0.0765 \pm 0.029$ & $0.0507 \pm 0.026$ \\
		\addlinespace[0.15cm]
		{GPPL} & $0.8942 \pm 0.001$ & $0.7213 \pm 0.004$ & $0.2047 \pm 0.010$ & $0.2074 \pm 0.010$\\
		{GPPL only embedding features} & $0.8224 \pm 0.004$ & $0.5687 \pm 0.009$ & $\mathbf{0.2333} \pm 0.022 $ & $\mathbf{0.2322} \pm 0.022 $ \\
		{GPPL only score features} & $0.9012 \pm 0.000$ & $\mathbf{0.7395} \pm 0.000$ & $0.1307 \pm 0.000$ & $0.1317 \pm 0.000$ \\
		\bottomrule
	\end{tabular}
	\caption[Effectiveness and Completeness Measurements]{Effectiveness and completeness on the test set of ACL-2018. The mean and standard deviation over five runs are given. "norm." refers to the citation counts normalized per track. "PRAUC" is area under the precision-recall curve.}
	\label{at:effect-test} 
\end{table*}
\begin{table*}[t]
	\centering
	\begin{tabular}{R{1.9cm}C{1.2cm}C{1.2cm}C{1.2cm}C{1.2cm}C{1.1cm}C{1.0cm}C{1.1cm}C{1.0cm}C{1.1cm}C{1.0cm}}
		\toprule
		& 
		\multicolumn{2}{C{2.5cm}}{\textbf{30\% noise}} &
		\multicolumn{2}{C{2.5cm}}{\textbf{60\% noise}} &
		\multicolumn{2}{C{2.5cm}}{\textbf{COMM-EQ}} & 
		\multicolumn{2}{C{2.5cm}}{\textbf{COMM-READ}} & 
		\multicolumn{2}{C{2.5cm}}{\textbf{COMM-CON}}\\
		\cmidrule(lr){2-3}\cmidrule(lr){4-5}\cmidrule(lr){6-7}\cmidrule(lr){8-9}\cmidrule(lr){10-11}
		& \textbf{AUROC} & $\rho$ \textbf{raw} & \textbf{AUROC} & $\rho$ \textbf{raw} & \textbf{AUROC} & $\rho$ \textbf{raw} & \textbf{AUROC} & $\rho$ \textbf{raw} & \textbf{AUROC} & $\rho$ \textbf{raw} \\
		\midrule
		MEAN-S-w & $\mathbf{0.889}$  & $0.119$ & $0.865$ & $0.078$ & $0.874$ & $0.149$  & $0.867$  & $0.089$ & $0.859$ & $0.144$  \\
		{DCON} & $0.820$  & $0.104$ & $0.794$ & $0.065$ &  $0.797$ & $-0.045$ & $0.795$ & $0.145$ & $0.809$ & $0.070$ \\
		GPPL & $0.886$ & $\mathbf{0.169}$ & $\mathbf{0.879}$ & $\mathbf{0.162}$ & $\mathbf{0.885}$  & $\mathbf{0.200}$ & $\mathbf{0.878}$ & $\mathbf{0.194}$ & $\mathbf{0.878}$  & $\mathbf{0.199}$  \\
		\bottomrule
	\end{tabular}
	\caption[Fairness]{Performance for the fairness scenarios. The noise scenarios refer to the case with $\sigma=1.0$ with $30\%$ and $60\%$ affected referees.}
	\label{at:fairness-test}
\end{table*}
\begin{figure}
	\centering
	\includegraphics[width=0.7\linewidth]{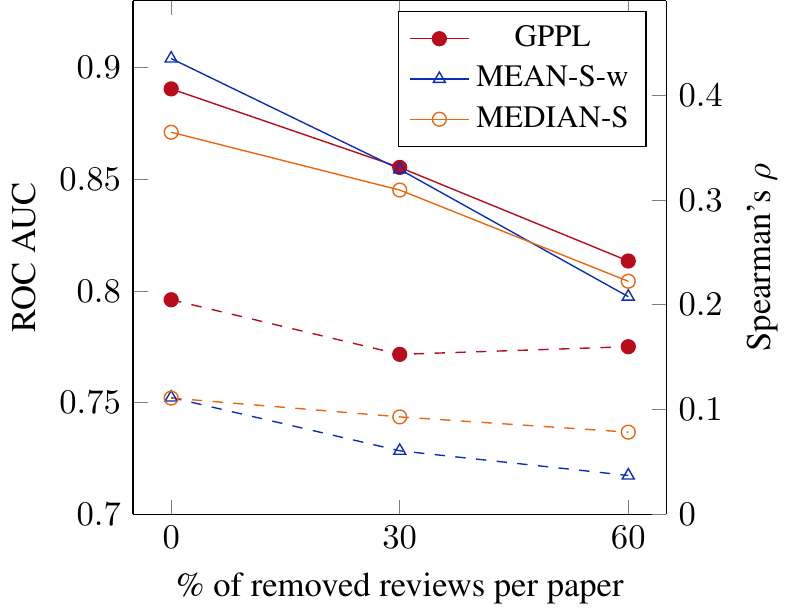}
	\caption{Performance on randomly removed reviews with at least one review per paper. Dashed lines refer to the right axis ($\rho$).}
\label{fig:efficiency}
\end{figure}
Our implementation of DCON is stopped early after $20$h of computation (8 CPUs, 16Gb of RAM).  NCON on average converges in $11$h. Our GPPL model using the final pre-computed feature set terminates on average in $4.5$min.
\paragraph{Feature Selection}
The best features on acceptance labels include score- and embedding-based features, but no discourse information. This configuration achieves $0.8463$ AUROC and $\rho=0.1930$ on citation counts. The performance drops to $0.7687$ AUROC without score-based features. Not surprisingly, review scores are central to predict acceptance labels, because they strongly influence decision making. To rank consistently with citation counts, embedding-based features are crucial: the best features include only  embeddings of the reviews' \textit{summary and contributions} sections and discourse features. This achieves $\rho = 0.2952$ and $0.6948$ AUROC. Discarding embedding-based features leads to a drastic drop of $\rho$ by $0.1049$.
This is reasonable, as impact is strongly linked to the contributions of a paper.
All further experiments use the features optimized on acceptance labels, because they offer the best performance trade-off on both gold standards.
\paragraph{Effectiveness and Completeness}
In Table \ref{at:effect-test} the effectiveness and completeness of our approach is compared to the baselines. As an ablation study, we also report the performance using only embedding-based and only score features.
While MEAN-S-w performs best on acceptance labels by AUROC, the difference to our model is close to zero ($-0.0099$). At the same time the performance gain of our model according to $\rho$ on the raw ($+184\%$) and normalized ($+65\%$) citation rankings is substantial. The consensus ranking baselines perform consistently worse. This shows the limitations of previous methods on real PR data.
The GPPL models using subsets of the best features confirm the importance of scores for acceptance labels and of the embedding-based features for citation counts. The low $\rho$ on citation counts of the MEAN-S-w also shows the difficulty of predicting future citation counts for human annotators putting the overall weak correlation into perspective. 
Finally, the equally high performance on track-wise normalized citation counts hints that our approach does not simply learn to favor topics with a broad audience, as this effect is mitigated in this ranking.
\paragraph{Fairness}
The results for simulating unreliable referees are consistent for added noise at levels $\sigma=0.75$ and $\sigma=1.0$. Hence, we only report on the setting with $\sigma=1.0$ on varying rates of unreliable referees.
For all algorithms except the consensus rankers, the ranking produced on the noisy reviews is highly consistent with their original rankings ($\rho > 0.9$). Table \ref{at:fairness-test} shows the AUROC and Spearman's $\rho$ on the test set for different ratios of affected referees. GPPL shows the smallest decay in performance for acceptance labels. On the citation count ranking, the performance of GPPL remains the highest, while the other baselines show a drastic drop for 60\% of noisy referees. Although adding noise only to scores favors approaches that rely on review texts, this suggests that the GPPL model is less affected by additional score noise.
For the three scenarios of commensuration bias, the performance of the best baselines and our approach is reported in Table \ref{at:fairness-test}. The GPPL model outperforms all other methods in all scenarios. Surprisingly, the COMM-EQ scenario leads to an improved performance on the citation count ranking for nearly all algorithms. Apparently, substituting the actual overall score by the average of aspect scores acts as a de-biasing approach. This suggests pre-processing of data samples and preference pairs might further improve performance.
\paragraph{Efficiency}
In the first scenario of efficiency evaluation ($\alpha=30\%$) on average $1.60$ reviews per paper and $1.93$ reviews per referee are left. For $\alpha=60\%$ of removed reviews, there are $1.01$ reviews per paper and $1.23$ reviews per referee. The consensus rankers are not applicable in both scenarios, as some papers are not included in a partial ranking of more than one element. Likewise, the GPPL model makes predictions on papers not seen during training.
As shown in Figure \ref{fig:efficiency}, the performance of all algorithms drops drastically. Our model deals slightly better with sparsity of reviews than the MEAN-S-w. The reduction of the number of reviews to increase efficiency appears contradictory to output quality for all approaches.
\section{Conclusion and Future Work}
In this paper, we demonstrated that preference learning is useful to assist acceptance decision making in peer review. We defined the Paper Ranking Problem and a novel generic evaluation framework to enable the empirical study of approaches in this field.  We showed that our GPPL-based method offers the best balanced performance on acceptance labels and citation counts, while being more robust against unreliable referees and added commensuration bias. Our experiments also highlighted the importance of both review texts and scores for ranking papers.
As future research directions specialized embeddings of review texts and their combination with paper embeddings proposed in scholarly document quality assessment are promising. Additionally, research on the transferability of our method to different peer review systems is essential once more datasets with complete peer review data become available.
\bibliographystyle{named}
\bibliography{peerpref}
\end{document}